\documentclass{article}

\usepackage[preprint]{neurips_data_2024}
\newcommand{\xhdr}[1]{\vspace{1em}\noindent{{\bf #1}}}

\usepackage[utf8]{inputenc} %
\usepackage[T1]{fontenc}    %
\usepackage{hyperref}       %
\usepackage{url}            %
\usepackage{booktabs}       %
\usepackage{amsfonts}       %
\usepackage{nicefrac}       %
\usepackage{microtype}      %
\usepackage{xcolor}         %
\usepackage{graphicx}
\usepackage{pifont}
\usepackage{rotating}
\usepackage{multirow}
\usepackage{appendix}
\usepackage{titletoc}

\usepackage{xspace}

\usepackage{listings}
\usepackage{xcolor}
\usepackage{caption}
\usepackage{float}
\newcommand\YAMLcolonstyle{\color{red}\mdseries}
\newcommand\YAMLkeystyle{\color{black}\bfseries}
\newcommand\YAMLvaluestyle{\color{blue}\mdseries}

\makeatletter

\newcommand\language@yaml{yaml}

\expandafter\expandafter\expandafter\lstdefinelanguage
\expandafter{\language@yaml}
{
  keywords={true,false,null,y,n},
  keywordstyle=\color{darkgray}\bfseries,
  basicstyle=\YAMLkeystyle,                                 %
  sensitive=false,
  comment=[l]{\#},
  morecomment=[s]{/*}{*/},
  commentstyle=\color{purple}\ttfamily,
  stringstyle=\YAMLvaluestyle\ttfamily,
  moredelim=[l][\color{orange}]{\&},
  moredelim=[l][\color{magenta}]{*},
  moredelim=**[il][\YAMLcolonstyle{:}\YAMLvaluestyle]{:},   %
  morestring=[b]',
  morestring=[b]",
  literate =    {---}{{\ProcessThreeDashes}}3
                {>}{{\textcolor{red}\textgreater}}1     
                {|}{{\textcolor{red}\textbar}}1 
                {\ -\ }{{\mdseries\ -\ }}3,
}

\lst@AddToHook{EveryLine}{\ifx\lst@language\language@yaml\YAMLkeystyle\fi}
\makeatother

\usepackage[]{roboto-mono}
\definecolor{textblue}{rgb}{.2,.2,.7}
\definecolor{textred}{rgb}{0.54,0,0}
\definecolor{textgreen}{rgb}{0,0.43,0}
\definecolor{codegreen}{rgb}{0,0.6,0}
\definecolor{codegray}{rgb}{0.5,0.5,0.5}
\definecolor{codepurple}{rgb}{0.58,0,0.82}
\definecolor{backcolour}{rgb}{0.92,0.92,0.92}

\usepackage{listings}

\lstdefinestyle{code}{
    language=Python, 
    numbers=left, 
    numberstyle=\tiny, 
    stepnumber=1,
    numbersep=5pt, 
    tabsize=4,
    basicstyle=\small\ttfamily,
    keywordstyle=\color{textblue},
    commentstyle=\color{textred},   
    stringstyle=\color{textgreen},
    frame=none,                    
    columns=fullflexible,
    keepspaces=true,
    xleftmargin=\parindent,
    showstringspaces=false
}

\lstdefinestyle{YAML}{
    language=yaml, 
    numbers=left, 
    numberstyle=\tiny, 
    stepnumber=1,
    numbersep=5pt, 
    tabsize=4,
    basicstyle=\ttfamily,
    frame=none,                    
    columns=fullflexible,
    keepspaces=true,
    xleftmargin=\parindent,
    showstringspaces=false
}

\lstdefinestyle{bash}{
    backgroundcolor=\color{backcolour},
    commentstyle=\color{textred},
    keywordstyle=\color{codepurple},
    numberstyle=\tiny\color{codegray},
    stringstyle=\color{codegreen},
    basicstyle=\ttfamily\footnotesize,
    breakatwhitespace=false,
    breaklines=true,
    captionpos=b,
    keepspaces=true,
    numbers=left,
    numbersep=5pt,
    showspaces=false,
    showstringspaces=false,
    showtabs=false,
    tabsize=2,
    language=bash
}

\title{MEDS-Tab: Automated tabularization and baseline methods for MEDS datasets}

\newcommand{\addrmit}{Massachusetts Institute of Technology, Cambridge, MA, USA}
\newcommand{\addmgh}{Massachusetts General Hospital, Boston, MA, USA}
\newcommand{\addrharvard}{Harvard Medical School, Boston, MA, USA}

\author{%
  Nassim Oufattole$^1$\thanks{Equal contribution}\\
  \textit{nassim@mit.edu} \\
  \And
  Teya Bergamaschi$^1$\footnotemark[1]\\
  \textit{teya@mit.edu} \\
  \And
  Aleksia Kolo$^1$ \\
  \textit{aleksiak@mit.edu } \\
  \And
  Hyewon Jeong$^1$ \\
  \textit{hyewonj@mit.edu} \\
  \And
  Hanna Gaggin$^{2}$\\
  \textit{hgaggin@mgh.harvard.edu} \\
  \And
  Collin Stultz$^{1,2,3}$\thanks{Corresponding author}\\
  \textit{cmstultz@csail.mit.edu} \\
  \And
  Matthew B.A. McDermott$^{3}$\footnotemark[2]\\
  \textit{matthew\_mcdermott@hms.harvard.edu} \\
  \AND
    \textnormal{$^1$ \addrmit} \\ 
    $^2$ \addmgh \\
    $^3$ \addrharvard
}

\begin{document}

\maketitle

\begin{abstract}
Effective, reliable, and scalable development of machine learning (ML) solutions for structured electronic health record (EHR) data requires the ability to reliably generate high-quality baseline models for diverse supervised learning tasks in an efficient and performant manner. Historically, producing such baseline models has been a largely manual effort--individual researchers would need to decide on the particular featurization and tabularization processes to apply to their individual raw, longitudinal data; and then train a supervised model over those data to produce a baseline result to compare novel methods against, all for just one task and one dataset. In this work, powered by complementary advances in core data standardization through the MEDS framework, we dramatically simplify and accelerate this process of tabularizing irregularly sampled time-series data, providing researchers the ability to automatically and scalably featurize and tabularize their longitudinal EHR data across tens of thousands of individual features, hundreds of millions of clinical events, and diverse windowing horizons and aggregation strategies, all before ultimately leveraging these tabular data to automatically produce high-caliber XGBoost baselines in a highly computationally efficient manner. This system scales to dramatically larger datasets than tabularization tools currently available to the community and enables researchers with any MEDS format dataset to immediately begin producing reliable and performant baseline prediction results on various tasks, with minimal human effort required. This system will greatly enhance the reliability, reproducibility, and ease of development of powerful ML solutions for health problems across diverse datasets and clinical settings.
\end{abstract}

\section{Introduction}

It is well established that tabular baseline methods, such as those produced by the XGBoost library~\cite{chen2016xgboost}, are highly competitive in comparison to neural network solutions, particularly in the spaces of tabular and structured, longitudinal medical data \cite{tabular_baselines, 8_pmlr-v219-labach23a, ebcl, omoplearn}. Currently, in the machine learning (ML) for healthcare space, researchers must produce these baseline comparison results by manually crafting their own heterogeneous pipelines to tabularize, featurize, and tune these methods on the diverse tasks of interest in medical AI. 

While this fact may seem a natural consequence of the prevalence of private datasets and unique data schemas in healthcare, it nevertheless causes significant problems for ML researchers in this space. Specifically, it is a notable waste of research time to functionally re-implement conceptually identical baseline pipelines across different datasets or tasks. Furthermore, it undermines the robustness and reproducibility of claims in ML for healthcare, as all comparisons against baselines must be interpreted as relative to the efficacy and level of appropriate tuning of the bespoke baseline pipeline used in the individual work being examined.

\begin{figure*}[t!]
  \centering
  \makebox[\linewidth][c]{\includegraphics[width=1\linewidth]{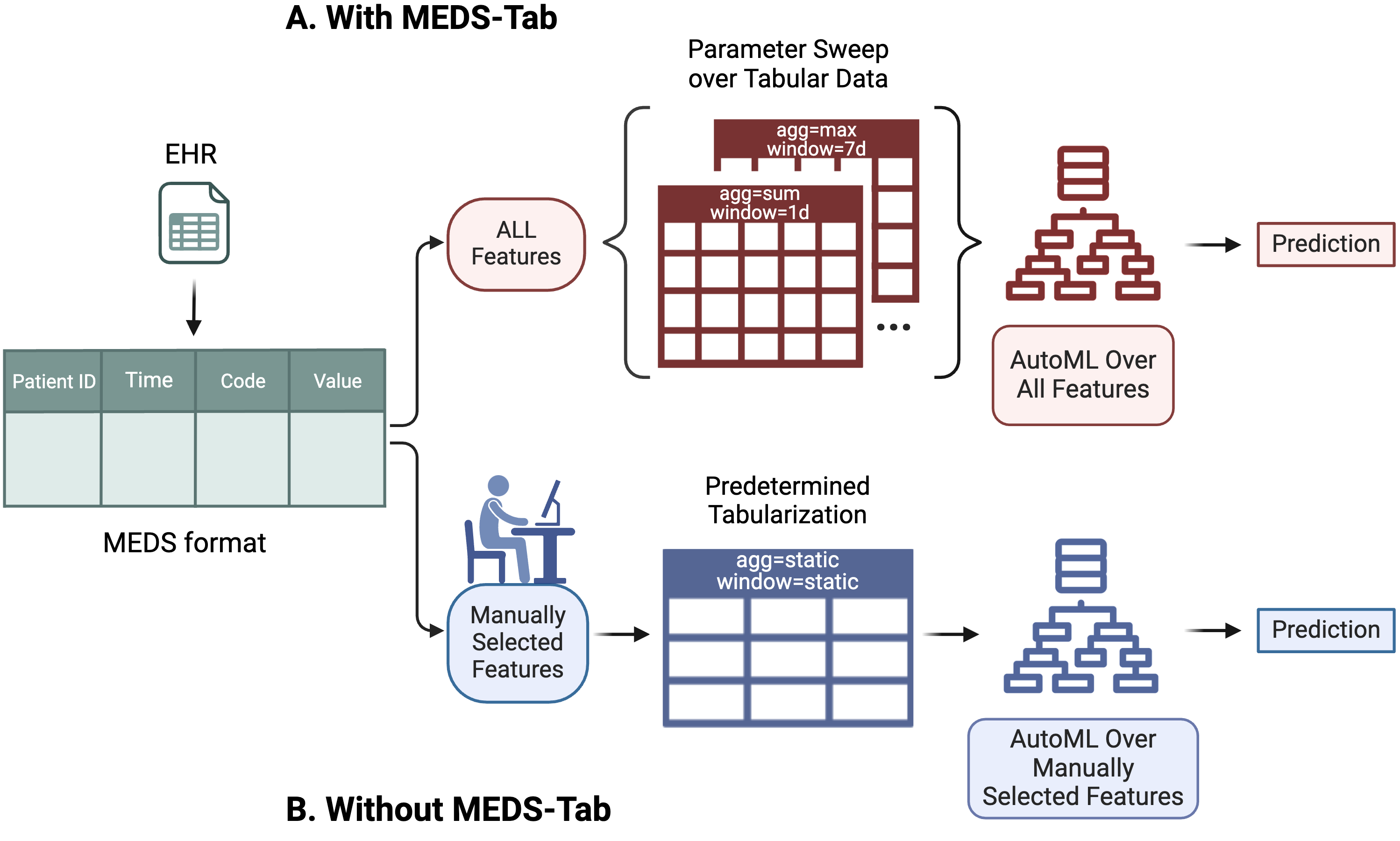}}
  \caption{MEDS-Tab: automated tabularization, data preparation with aggregation and windowing.}
  \label{fig:Concept}
\end{figure*}

To address these problems, the medical ML community is in desperate need of easy-to-use tools that can consistently produce competitive baselines across diverse EHR datasets and tasks. In this work, we provide such a tool by releasing MEDS-Tab: a tabularization and XGBoost AutoML \cite{singh2022automated, rashidi2021machine} pipeline for longitudinal medical data (Figure \ref{fig:Concept}). MEDS-Tab leverages the recently developed, minimal, easy-to-use Medical Event Data Standard (MEDS) \cite{MEDS} schema to standardize structured electronic health record (EHR) data to a consistent schema from which baselines can be reliably produced across arbitrary tasks and settings. MEDS-Tab scales to extremely large health datasets with hundreds of millions of clinical events and tens of thousands of unique medical codes, and it significantly reduces the engineering burden for producing competitive baselines.

In sum, we introduce a consistent and generalizable tool that (1) tabularizes longitudinal, structured, event-stream medical data in an efficient, highly flexible, and dataset-agnostic manner, and then (2) leverages that tabularized data using AutoML tools to tune high-performance tree-based ML methods on large-scale medical datasets for arbitrary downstream tasks. In concert with MEDS and its ecosystem, this tool enables researchers to \emph{reliably profile baseline performance for both novel and existing downstream tasks across diverse EHR datasets or publications}. It encoura that results are reproducible, trustworthy, and easily communicated with minimal human effort. This advancement significantly reduces the burden on researchers by facilitating their ability to work with both existing and new datasets, communicate findings in scientific publications in a reproducible manner, reproduce findings from other researchers, and develop performant baseline models within a controllable computational budget.

The rest of this paper is structured as follows: First, in Section \ref{sec:problem}, we describe the problem of tabularization and baseline model generation over structured, longitudinal medical data in more detail. Then, we present our innovative approach to overcome these issues in Section \ref{sec:method}. Finally, we discuss the broader implications and the conclusion of our findings in Sections \ref{sec:discussion} and \ref{sec:conclusion}.

\section{Problem Description} \label{sec:problem}

This work addresses the challenges of generating a baseline model that leverages all available time-series observations from the EHR given a medical dataset and labels for some prediction task of interest. We restrict ourselves to decision tree models (specifically using XGBoost \citep{chen2016xgboost}) as these models are extremely performant and widely used. We need this baseline to be reproducible on any medical dataset, so users can confidently and reproducibly run the same baseline, regardless of the nuances of their EHR dataset. Additionally, we need these baselines to scale to large medical datasets. This problem contains two main steps: converting a raw EHR dataset into a model ingestible format and training and tuning the decision tree. We precisely define these stages below:

\xhdr{Step 1: Tabularization}
Structured medical time series data, while often referred to as "tabular," is not typically in a format directly usable by tabular models like XGBoost. These models require data where each column is a unique feature and each row represents a single instance.

Tabularization is the process of converting data into a format suitable for decision tree models. It consists of summarizing all time-series data for a subject up to an event-time, into a fixed-size tabular feature vector. These feature vectors are then paired with corresponding prediction labels and can be fed to a decision tree for training and evaluation. This process proceeds as follows:

First, Users select a set of aggregation functions (e.g., sum, count, average) and window sizes. Then, for each unique combination of aggregation function and window size 
\begin{enumerate}
    \item The time series data is filtered to include only the data within the specified window up to the event time.
    \item The chosen aggregation function is applied independently to each code (feature) within this filtered data.
\end{enumerate}

This process transforms the data into a "wide" tabular format where:

Each column represents a unique combination of code, time window, and aggregation function. Each row represents an event, with the aggregated values for each code-window-function combination. This approach allows for flexible summarization of time-based features, capturing different aspects of the data (e.g., recent trends, long-term patterns) through various user-selected aggregation methods and time scales.

\xhdr{Step 2: Training a baseline model}
The process of training a baseline model begins with data selection from the tabularized dataset created in Step 1. This involves identifying and extracting the most recent feature vector for each subject prior to the designated prediction time and then matching these vectors with the corresponding labels provided by the user. Once this selection is complete, the task shifts to efficient model training. A baseline model, such as a decision tree, is trained using these selected feature vectors and their associated labels. An AutoML tool (such as Optuna \citep{akiba2019optuna}) can be used for tuning hyperparameters.

\subsection{Challenges}
\xhdr{Data Processing}
Data tabularization is commonly broken into two steps: transformation to long form, event-stream data, and conversion from long form to wide "tabular" data. While we acknowledge the challenge of processing data to long form, several existing tools have been developed to address this issue. For example:

\begin{itemize}
    \item \href{https://github.com/mmcdermott/MEDS_transforms}{MEDS\_Transforms}
 is a tool designed to convert raw EHR data, often stored in multiple CSV files, into the Medical Event Data Standard (MEDS) format.
 \item \href{https://github.com/Medical-Event-Data-Standard/meds_etl}{meds\_etl} is another tool that specializes in converting OMOP v5 data into the MEDS format.
\end{itemize}

These tools demonstrate that while the process of transforming diverse EHR data formats into a standardized long-form representation is complex, it is a challenge that has been addressed by the research community. By leveraging these existing solutions, researchers and practitioners can more easily overcome the initial hurdle of data preprocessing, allowing them to focus on subsequent steps in the analysis pipeline.

\xhdr{Scalability of Tabularization}
Na\"ively attempting to turn EHR data, even long-form data, into tabular features can result in a serious computational hurdle. Namely, realizing medical data across a unified vocabulary of categorical "codes" results in datasets with extremely large numbers of codes (e.g., tens of thousands or more). This means that the creation of this wide-form matrix poses multiple computational challenges \cite{borisov2022deep}. First, the transformation requires very large amounts of memory, which in turn can impose a prohibitively high barrier to training and deploying these models as computational budgets vary widely across different settings. Furthermore, the wall time required to generate these features can become prohibitively long as the code count or number of samples increases. 

\xhdr{Model Reproducibility}
Reproducibility challenges are not merely theoretical but are evidenced in recent literature. In a brief survey of three recent conferences on ML for healthcare, Machine Learning for Healthcare Conference \cite{1_MLHC2023}, Machine Learning for Health \cite{2_pmlr-v225-hegselmann23a}, and Conference on Health, Inference, and Learning \cite{3_pmlr-v248-pollard24a}, we found 12 papers \cite{4_pmlr-v219-elhussein23a, 5_pmlr-v219-ho23a, 6_pmlr-v225-king23a, 7_pmlr-v225-kuznetsova23a, 8_pmlr-v219-labach23a, 9_pmlr-v225-noroozizadeh23a, 10_pmlr-v225-ren23a, 11_pmlr-v225-van-ness23a, 12_pmlr-v248-xu24a, 13_pmlr-v225-xu23a, 14_pmlr-v248-yeche24a, 15_pmlr-v219-zhang23a} using longitudinal EHR data. Of those 12 papers, 83\% of papers \cite{5_pmlr-v219-ho23a, 6_pmlr-v225-king23a, 7_pmlr-v225-kuznetsova23a, 8_pmlr-v219-labach23a, 9_pmlr-v225-noroozizadeh23a, 10_pmlr-v225-ren23a, 11_pmlr-v225-van-ness23a, 12_pmlr-v248-xu24a, 14_pmlr-v248-yeche24a, 15_pmlr-v219-zhang23a} included a tabular baseline when reporting task specific results, and all of these papers use manual feature selection. This manual feature selection process also compounds the major reproducibility challenges present in machine learning for health, especially because approximately 58\% of studies do not share their data processing code, rendering the details of these model training recipes obscured from the community. Additionally, manual feature selection reduces the extensibility to new datasets. Our tool addresses these challenges by enabling researchers to fit a well-tested and clearly communicated feature extraction and tabular baseline via a method that can be methodologically transposed and deployed on any MEDS dataset, thus enhancing reproducibility and standardization in the field. Moreover, our approach scales up to allow the inclusion of all features, overcoming the limitations of manual feature selection and potentially capturing more comprehensive patterns in the data.

\xhdr{Scalability of Model Training}
A significant challenge in decision tree training is the efficient loading of large-scale data. Implementing this process inefficiently can result in high loading times, dramatically increasing overall model training duration. To address this, it's crucial to develop techniques for efficient data loading, especially when dealing with datasets too large to fit in memory. This optimization is particularly important for AutoML pipelines that run multiple hyperparameter trials in parallel, which require the ability to efficiently load and process subsets of the data (concurrently for multiple models) for training and evaluation.

The AutoML pipeline must be designed to perform effective tuning of model hyperparameters and feature subsets without incurring excessive computational overhead. This involves striking a balance between leveraging the vast amounts of tabularized medical data and maintaining practical computational efficiency. By optimizing these aspects, the training process can efficiently handle large-scale data, enabling more effective model development and evaluation in the context of medical predictive tasks.

\section{MEDS-Tab: Tabularization and baseline AutoML for MEDS dataset.} \label{sec:method}

We present MEDS-Tab as a robust baselining solution specifically designed to overcome the complex computational challenges posed by tabularizing EHR data at a large scale and to facilitate the efficient use of AutoML pipelines for arbitrary supervised tasks on these same large datasets, all while minimizing user effort. MEDS-Tab accomplishes this by explicitly managing and optimizing the use of computational resources through several strategic implementations and by providing a user-friendly command line interface for easy deployment. 

\subsection{MEDS-Tab Implementation}

\subsubsection{Tabularization}
MEDS-Tab expects input data to be stored in a long format. From this format, the challenge of tabularization becomes how to summarize a subject's data until the prediction time into a fixed-size view where every column is a feature.
The natural way to do this is to break the problem down into two steps: first, converting the data from the "long" form where each row contains a single observation, specifying for which code any observation applies, to a "wide" form where all unique codes of the data are realized as different columns and rows corresponding only to unique subject events in time; and, second, aggregating this wide format data frame over varying historical windows to produce a fixed-size, non-temporal summary of the subject's history as of a given prediction time.

MEDS-Tab employs multiple methods to optimize this tabularization step: 
\paragraph{Sparse Tabular Representation} MEDS-Tab employs a sparse data format for storage and computation. This approach significantly reduces the memory footprint by only storing non-zero elements, which is particularly effective given the sparse nature of medical time-series datasets, and speeds up the computation of aggregations over varying window sizes. The constructed tabular features describe subject records over arbitrary time windows. The system supports a wide variety of aggregation methods and can handle any window size for analysis. It is capable of tabularizing four types of data: static codes, static numerical values, time-series codes, and time-series numerical values, facilitating comprehensive data structuring and analysis.

\paragraph{Data Sharding} To scale to very large healthcare datasets, MEDS-Tab uses a sharded data model, where data is chunked into smaller subsets of subjects. This allows the larger processes of tabularization and model training to be separated over smaller, more manageable sets of data. For tabularization, each shard can be processed independently, enhancing scalability and enabling parallelization both locally and even across multi-node slurm clusters. 

\paragraph{Polars computation} To tabularize data, MEDS-Tab iterates through combinations of window sizes and aggregation methods to generate feature vectors for unique events, \textit{subject\_id} $\times$ \textit{timestamp}, on a per shard level and uses sparse matrix formats to efficiently handle the computational and storage demands. Data is pre-sorted by subject ids and time stamps, and polars is used to efficiently pre-compute rolling indices for the rolling window aggregations. By separating events into reasonably sized shards and leveraging the low memory cost of sparse matrices, this computation becomes incredibly efficient and highly parallelizable, significantly reducing the required wall time for tabularizing the data (see the Appendix for computational overhead comparison to other methods). 

\subsubsection{Model Training on Large Datasets}
For datasets that exceed typical memory capacities, MEDS-Tab supports extended memory training, facilitating training on datasets at scales too large to be fully loaded onto memory. This is achieved by efficiently loading data shards from disk sequentially during model training, thus trading off latency when loading data in for reduced RAM usage. A naïve implementation of this shared data loading would quickly become impractical due to ballooning wall time, even for in-memory training. MEDS-Tab employs multiple design choices, in addition to sparse matrices, to combat this problem. 

\paragraph{Task Specific Data Caching and Loading} Tabularization is conducted over the entire dataset; however, for any given task, only a small subset of events will be relevant. 
An optimal implementation would only ever load the relevant events. To accomplish this, MEDS-Tab aligns task-specific labels with the nearest prior event in the tabularized data,
and discards all unmatched events. In doing this, only relevant events, represented as rows in the tabularized dataset, are kept. For example, during tabularization, a subject \textit{X} may have events \textit{1}-\textit{50}; however, for the task of predicting hospital readmission, predictions should only be made at the discharge events, which in this example may be events \textit{4, 10, 36}. Therefore, out of the original 50 rows occupied by subject \textit{X} only those three event rows (\textit{4, 10, 36}) would be kept in the task-specific cache files. This improves data loading efficiency by eliminating the need for on-the-fly and repeated row selection during training and by caching smaller files in which all rows are relevant for training. 

\paragraph{Extended Memory and CPU Optimization} MEDS-Tab pre-caches task-specific data shards, ensuring only task-relevant event rows are ever loaded during training, and the method carefully optimizes and minimizes necessary data manipulation steps during data loading to limit this potential bottleneck.

\paragraph{Flexible AutoML Pipeline} MEDS-Tab includes a flexible AutoML pipeline powered by Optuna, which automates the tuning of model hyperparameters and featurization options including data aggregation methods, rolling window sizes, and the selection of relevant medical codes. 
While our system incorporates basic AutoML capabilities, its primary innovation lies in the preparation and management of data for these algorithms, facilitating extensive experimentation with different featurization strategies to optimize predictive modeling tasks.

\subsection{Command Line Interface}

The design choices above have been realized into an easy-to-use command line interface. Starting with a dataset in MEDS format and labels for a prediction task of interest, the following five commands are all that is needed to tabularize the data and train a supported model:

\xhdr{Data Description (\textit{`meds-tab-describe'})} Analyzes MEDS data shards to compute code frequencies, categorizing them into time-series codes, static codes, and their numerical variants. Results are cached in a \textit{`code\_metadata.parquet`} file. These frequencies can optionally used to filter down to codes with a minimum frequency via adding the argument \textit{`tabularization.min\_code\_inclusion\_frequency=X`} to the following steps.

\xhdr{Static Data Tabularization (\textit{`meds-tab-tabularize-static'})} Transforms static subject data into a tabular format, creating feature vectors for each subject at each timestamp based on specified code frequencies and aggregation methods. Generally, the number of static features is not very large in EHR datasets, so a sparse matrix is not necessary; however, we currently store these as sparse matrices so they can be quickly concatenated during model training with sparse matrices generated during the time-series data tabularization step.
\begin{figure*}[t]
  \centering
  \makebox[\linewidth][c]{\includegraphics[width=1\linewidth]{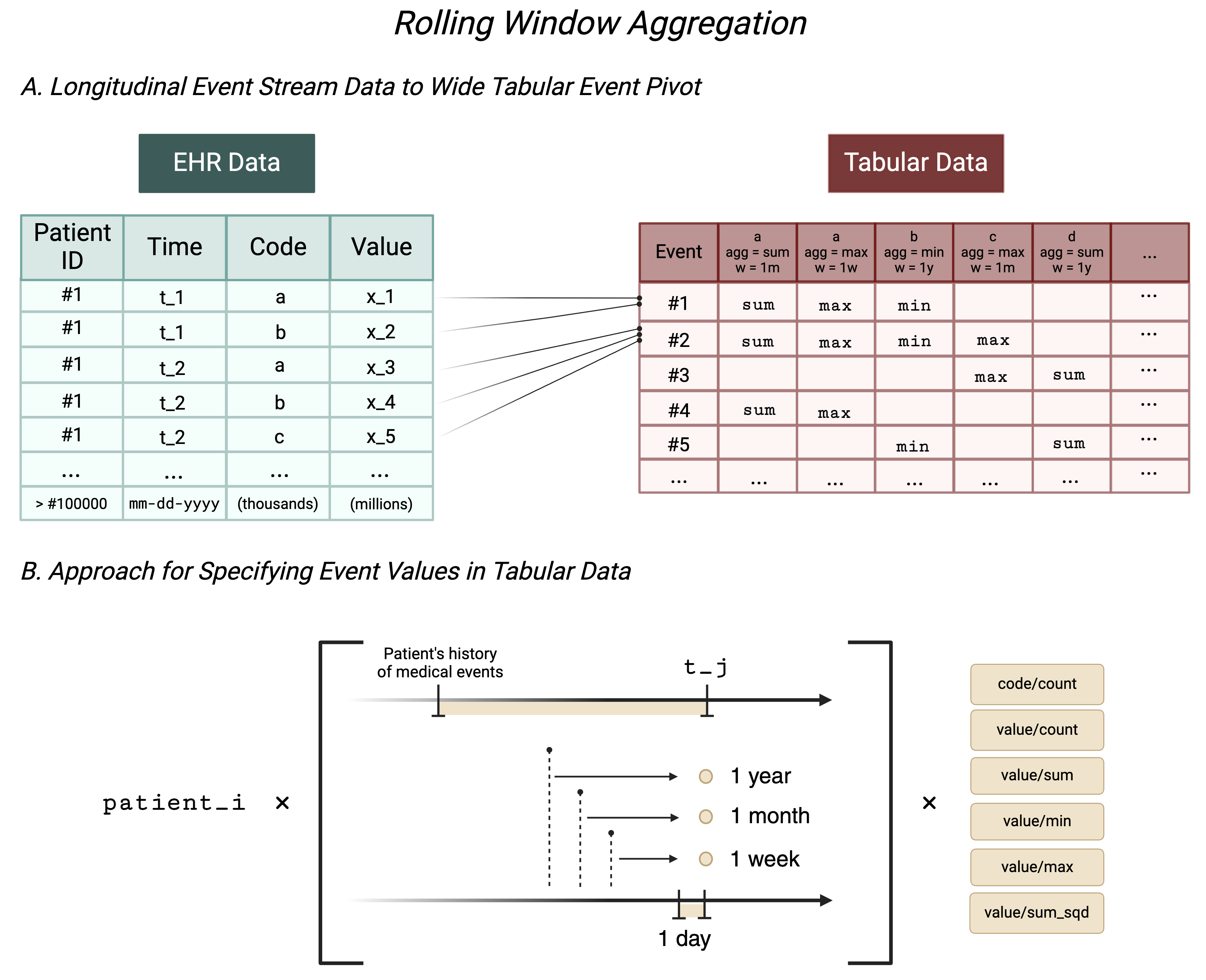}}
  \caption{\textbf{EHR Data Featurization.} \textit{Tabularization} process for summarizing temporal data into feature vectors where the features are aggregated over a multitude of lookback windows which are concatenated into summarized features.}
  \label{fig:pivot_operation}
\end{figure*}

\xhdr{Time-Series Data Tabularization (\textit{`meds-tab-tabularize-time-series'})} Generates feature vectors by aggregating subject time-series data across various window sizes and aggregation methods (as shown in Figure \ref{fig:pivot_operation}), utilizing sparse matrix formats for efficiency. More specifically, this step iterates through combinations of window sizes and aggregation methods to generate feature vectors for every unique time a subject has a measurement at (i.e. event times), subject\_id $\times$ timestamp, on a per shard level and uses sparse matrix formats to efficiently handle the computational and storage demands. Data is pre-sorted by subject\_ids and time stamps, and polars is used to efficiently pre-compute rolling indices for the rolling window aggregations. By separating events into reasonably sized shards and leveraging the low memory cost of sparse matrices, this computation becomes incredibly efficient and highly parallelizable, significantly reducing the required wall time for tabularizing the data (see the Appendix for computational overhead comparison to other methods). In this one command, the tabularization of time-series data is completed and the resulting tabularized data is ready for use either in the MEDS-Tab training pipeline or elsewhere.

\xhdr{Task-Specific Label Alignment (\textit{`meds-tab-cache-task'})} A prediction task is provided by the user through a table with columns subject\_id, timestamp, and label, following the same sharding and file structure as the original sharded dataset. To perform XGBoost training, for each shard, we need to align these task-specific labels with the closest feature vector (from tabularization) with an event time before the prediction timestamp (otherwise there will be data leakage as the feature vector includes information after the prediction time). It is precisely this alignment that is performed in this stage, and sparse \& sharded matrices that are filtered and aligned to these labels are generated and stored in this stage. 

\xhdr{Model Training (\textit{`meds-tab-model'})} Trains an XGBoost or one of the supported SciKit-Learn \cite{pedregosa2011scikit} classifiers using the prepared data, allowing for an Optuna AutoML sweep over different combinations of window sizes and aggregation methods. For datasets that do not fit on memory, this stage additionally supports training via loading only one shard at a time via adding the command line flag \textit{`model\_params.iterator.keep\_data\_in\_memory=False'} for any model that supports partial fit.

Further details on MEDS-Tab CLI can be found in MEDS-Tab's publicly available documentation: \url{https://meds-tab.readthedocs.io/en/latest/}.

\subsection{Data Processing and Feature Selection}
While MEDS-Tab is specifically designed to leverage all available data with little required processing, there are instances in which further processing, such as normalization, or feature selection may be useful. As such, MEDS-Tab supports various data processing and feature selection methods that may be useful for leveraging various SciKit-Learn models.
\subsubsection{Data Processing}
Tree-based methods, such as XGBoost, are insensitive to normalization \cite{de2023choice, hastie2009elements} and generally do not achieve higher performance from data imputation of missing values \cite{rusdah2020xgboost, aydin2021performance}. In fact, XGBoost natively handles learning what decision to make when encountering missing data \cite{chen2016xgboost}. As a result, our tool by default does no further preprocessing to normalize or impute the tabularized data. 

However, other supported models, such as \textit{kneighbors\_classifier}, \textit{logistic\_regression}, and \textit{sgd\_classifier} do not universally handle missingness as gracefully. Therefore, MEDS-Tab supports (mean \textit{mean\_imputer}, median \textit{median\_imputer}, and mode \textit{mode\_imputer}) imputation. It is important to note that imputation methods require making the data dense as missing values are filled with imputed values, which can drastically affect computation performance. 
\subsubsection{Feature Selection}
While the goal of MEDS-Tab is to facilitate training on datasets with many features, it can be useful to impose restrictions on the available features. As such, MEDS-Tab supports 5 feature selection methods:
\begin{enumerate}
    \item Allowed codes -- \textit{tabularization.allowed\_codes} -- manually select which codes will be included as features (conducted on codes \textit{before} aggregation $\times$ window size featurization).
    \item Codes with X prevalence -- \textit{tabularization.min\_code\_inclusion\_count} -- only include codes that were measured at least X times across the dataset (conducted on codes \textit{before} aggregation $\times$ window size featurization).
    \item Top N most common codes -- \textit{tabularization.max\_included\_codes} -- (conducted on codes \textit{before} aggregation $\times$ window size featurization).
    \item Top R approximate correlation -- \textit{tabularization.max\_by\_correlation} -- include only the top R features with the highest approximate correlation to the target label (conducted on features \textit{after} aggregation $\times$ window size featurization).
    \item Features with at least C approximate correlation with the target label -- \textit{tabularization.min\_correlation} -- only include features with at least $|C|$ approximate correlation with the target label (conducted on features \textit{after} aggregation x window size featurization).
\end{enumerate}

\section{Discussion} \label{sec:discussion}

\subsection{Related Works}

Recent advancements in machine learning for medical tabular time-series datasets have established a solid foundation for significant progress in healthcare analytics. Notably, works that provide large-scale EHR and trial datasets \cite{johnson2020mimic, pollard2018eicu, fu2022hint} have enabled researchers to develop models for critical applications such as patient outcome prediction \cite{pang2022establishment, sun2023prediction} and treatment effect estimation \cite{sun2022effects}. These efforts underscore the complex nature of medical data which is often high-dimensional and sparsely measured, presenting unique challenges for modeling \cite{jensen2012mining, weiskopf2013defining, wells2013strategies}. In addressing these complexities, the existing body of work can be broadly categorized into three main areas: the handling and tabularization of irregular temporal data, the automation of modeling processes through techniques such as AutoML \cite{singh2022automated, rashidi2021machine}, and the streamlining of pipelines to enhance efficiency and robustness.

\subsubsection{Tabularization of Time-Series Data}
The process of converting irregularly sampled time-series data into a structured tabular format is essential for the effective application of machine learning models. Tools like sktime \cite{loning2019sktime}, tsfresh \cite{christ2018time}, Clairvoyance \cite{jarrett2023clairvoyance}, and TemporAI \cite{saveliev2023temporai} have significantly advanced this area by offering robust frameworks for feature extraction and transformation. These tools enable the tabularization of temporal data by aggregating features across various time windows and handling missing values, which is critical for maintaining the integrity and predictive power of the resulting models. For example, tsfresh automates the extraction of relevant features from time-series data, while sktime provides a unified framework for time-series analysis that integrates seamlessly with existing machine learning libraries. However, tsfresh states that the memory consumption of the parallelized calculations can be high, which can make the usage of a high number of processes on machines with low memory infeasible, and methods in sktime only support data with equal length series and no missing values. In the healthcare domain, Clairvoyance which has been superseded by TemporAI offers toolkits to handle tabularization, but both systems use data containers like \textit{pandas.DataFrame} or \textit{numpy.ndarray} which do not handle the sparse nature of EHR data.

\subsubsection{AutoML} The automation of machine learning workflows, particularly through AutoML, has greatly simplified the process of model development, making it accessible to non-experts and reducing the time required to achieve competitive results. Tools such as AutoGluon \cite{erickson2020autogluon}, Optuna, and hyperopt \cite{bergstra2015hyperopt} exemplify the power of AutoML. These systems automate elements of algorithm selection, hyperparameter tuning, and model evaluation, thereby optimizing the modeling process. For instance, AutoGluon offers a comprehensive AutoML framework automating the end-to-end process of model selection, hyperparameter tuning, and ensembling across various tasks, including classification and regression on medical datasets. While AutoGluon is designed to be easy to use and powerful, the automation of complex processes like model stacking, hyperparameter optimization, and ensembling can lead to significant demands on CPU, memory, and disk space, which can make it less suitable for environments with limited computational resources or when quick, lightweight model training is needed. Optuna has emerged as a powerful tool for hyperparameter optimization, offering both an efficient and flexible framework that enables fine-tuning of models to achieve optimal performance with minimal manual intervention.

\subsubsection{Integrated Pipelines}
While the contributions in the aforementioned areas are foundational, they often exist as isolated solutions that would benefit immensely from integration into a unified framework that ensures reproducibility and robustness across different datasets. Integrated pipelines such as CaTabRa \cite{maletzky2023catabra}, Cardea \cite{alnegheimish2020cardea}, Clairvoyance, and TemporAI aim to bridge this gap by combining the strengths of tabularization and AutoML within a cohesive framework. 

CaTabRa offers a largely automated workflow that includes model training, evaluation, explanation, and out-of-distribution (OOD) detection. The system integrates multiple established frameworks and libraries, such as auto-sklearn \cite{feurer2020auto}, into a coherent package that allows users to quickly gain insights from their data. However, CaTabRa’s approach of “automate what can reasonably be automated” means that more complex tasks may still require significant user intervention and expertise, particularly in the definition and extraction of meaningful target labels for supervised learning. Cardea is an open-source framework tailored specifically for electronic health records (EHR) data, automating the entire machine learning process from data cleaning to model deployment. It leverages tools like ML-Bazaar \cite{smith2020machine} and hyperopt for feature engineering and hyperparameter optimization. While Cardea provides a robust solution for EHR data, its general applicability is limited by the predefined problem definitions it supports. Clairvoyance developed as a pipeline toolkit for medical time series, focuses on facilitating the processing and modeling of time-series data. It integrates various preprocessing, feature extraction, and modeling components, along with hyperparameter optimization. However, Clairvoyance has been critiqued for its insufficient modularity, lack of robust testing, and limited support from the community. These issues have led to its partial supersession by TemporAI, which addresses these shortcomings by offering a more modular and community-supported framework. TemporAI  is an open-source Python library designed for machine learning tasks involving temporal data, particularly in the medical domain. It supports a variety of data modalities—time series, static, and event—and provides tools for prediction, causal inference, time-to-event analysis, and model interpretability. TemporAI's strengths lie in its comprehensive support for different data modalities and its integration of both deep learning and traditional algorithms. However, like its predecessor, given the data container choice of \textit{pandas.DataFrame} requires significant computational resources.

\subsubsection{Further Considerations} Additional contributions include encoding techniques \cite{tipirneni2022self} and robust modeling approaches like self-supervised learning \cite{ebcl, li2020learning} and multi-task learning \cite{harutyunyan2019multitask, nguyen2021clinical,mcdermott2020comprehensive}, all of which aim to manage the irregular sampling of datasets effectively. Moreover, the field has seen significant improvements in imputation strategies to address data missingness \cite{li2021imputation, jarrett2022hyperimpute}, enhancing the overall quality and usability of medical datasets. These collective efforts reflect a vibrant ongoing endeavor to refine data processing and analysis techniques in healthcare. In terms of toolkits that offer a wide range of machine learning architectures, PyHealth \cite{zhao2021pyhealth} provides access to over 30 state-of-the-art models, making it a robust choice for healthcare predictive modeling; however, its focus on providing a broad array of models rather than automated processes means that users may need significant expertise to effectively utilize and integrate these models into complex workflows.

MEDS-Tab emerges as part of this broader narrative, seeking not only to contribute to the existing efforts in the field but also to synthesize these disparate advances into a scalable and efficient tool tailored specifically for medical datasets. While tools like TemporAI, CaTabRa, and Cardea provide comprehensive systems for predictive modeling and data analysis, their reliance on dense data representations often leads to scalability and performance bottlenecks. The magnitude of these dense matrices makes featurizing across a wide range of window sizes and aggregations intractable \cite{zhou2017learning}. For example, just on the public dataset MIMIC-IV \cite{mimiciv}, there are over 4 hundred million unique events and around 30,000 features. Assuming 32-bit precision, a naïve extraction approach using past AutoML tabularization pipelines would require at least 48 terabytes of RAM. Datasets of this scale (and larger) are increasingly common, and the RAM requirements render the existing methods non-viable. Consequently, these tools may require significant feature reduction through selection algorithms, potentially reducing the performance of the predictive models generated. By addressing these challenges, MEDS-Tab aims to enable more reproducible and accessible model benchmarking on medical datasets.

\subsection{Case Studies and Performance Comparisons}

To illustrate the practical application of our pipeline and establish robust baselines within the medical ML community, we report competitive AUCs for a range of well-known clinical tasks, including readmission and mortality predictions on MIMIC-IV and Philips eICU. We have highlighted some of these results in Table \ref{tab:combined_xgboost}. Additionally, we conducted comparisons against established solutions such as CaTabRa and tsfresh, which highlighted our pipeline’s enhanced scalability and efficiency. Our approach significantly reduces memory usage and storage requirements while demonstrating substantially lower wall times. This efficiency is achieved by leveraging the embarrassingly parallel nature of the tabularization problem.
\begin{table}[htbp]
\normalsize
\caption{\textbf{XGBoost AUC and Dataset Size for MIMIC-IV and eICU Tasks:} This table illustrates the scalability of the XGBoost model across various clinical prediction tasks in the MIMIC-IV and eICU datasets, emphasizing the quick establishment of baseline performances enabled by MEDS-Tab. It includes subject and event counts to demonstrate the data scale and model applicability across a diverse set of clinical tasks.}
\label{tab:combined_xgboost}
\centering
\begin{tabular}{lllrrr}
\toprule
prediction timestamp & Dataset & Task & AUC & Subjects (k) & Events (k) \\
\midrule
\multirow{3}{*}{Discharge} & \multirow{3}{*}{MIMIC-IV} & Post-discharge 30 day Mortality & 0.935 & 149 & 356 \\
                           &  & Post-discharge 1 year Mortality & 0.898 & 149 & 356 \\
                           &  & 30 day Readmission & 0.708 & 17 & 378 \\
\cmidrule(r){1-6}
\multirow{7}{*}{Admission + 24 hr} & \multirow{4}{*}{MIMIC-IV} & In ICU Mortality & 0.661 & 8 & 23 \\
                                   &  & In Hospital Mortality & 0.812 & 51 & 339 \\
                                   &  & LOS in ICU > 3 days & 0.946 & 43 & 61 \\
                                   &  & LOS in Hospital > 3 days & 0.943 & 152 & 360 \\
\cmidrule(r){2-6}
                                   & \multirow{3}{*}{eICU} & In Hospital Mortality & 0.855 & 4 & 4 \\
                                   & & LOS in ICU > 3 days & 0.783 & 13 & 14 \\
                                   &  & LOS in Hospital > 3 days & 0.864 & 100 & 100 \\
\cmidrule(r){1-6}
\multirow{7}{*}{Admission + 48 hr} & \multirow{4}{*}{MIMIC-IV} & In ICU Mortality & 0.673 & 7 & 21 \\
                                   &  & In Hospital Mortality & 0.810 & 47 & 348 \\
                                   &  & LOS in ICU > 3 days & 0.967 & 43 & 61 \\
                                   &  & LOS in Hospital > 3 days & 0.945 & 152 & 359 \\
\cmidrule(r){2-6}
                                   & \multirow{3}{*}{eICU} & In Hospital Mortality & 0.570 & 2 & 2 \\
                                   &  & LOS in ICU > 3 days & 0.757 & 13 & 14 \\
                                   &  & LOS in Hospital > 3 days & 0.895 & 100 & 100 \\
\bottomrule
\end{tabular}
\end{table}
\begin{table}[htbp]
\normalsize
\caption{\textbf{Comparative Performance on eICU and MIMIC-IV Datasets:} This table demonstrates MEDS-Tab's efficient processing of eICU and MIMIC-IV datasets on high-performance hardware (2 x AMD EPYC 7713 CPUs, 1024GB RAM), with a 10-minute time limit per run. Wall times (mm:ss) show the processing duration for all subjects. We compute only code/count aggregation over the full subject history up to every event time, which involves counting the occurrences of each code (e.g., diagnoses, procedures) for each subject. This aggregation is the most computationally intensive task, effectively stress-testing all methods. Memory usage scales moderately with dataset size. MEDS-Tab outperforms tsfresh and CaTabRa, which fail to complete tabularization for larger datasets within the time limit, demonstrating superior scalability. The best results for each metric are in bold.}
\label{tab:combined_performance}
\centering
\begin{tabular}{lrrrrrl}
\toprule
Dataset & Subjects & Wall Time & Avg Memory (MB) & Peak Memory (MB) & Method \\
\midrule
\multirow{2}{*}{eICU} & 100 & \textbf{0:39} & \textbf{5,271} & \textbf{14,791} & MEDS-Tab \\
\cmidrule(r){2-6}
 & 500 & \textbf{3:04} & \textbf{8,335} & \textbf{15,102} & MEDS-Tab \\
\cmidrule(r){1-6}
\multirow{6}{*}{MIMIC-IV} & \multirow{3}{*}{10} & \textbf{0:02} & \textbf{423} & \textbf{943} & MEDS-Tab \\
 & & 1:41 & 84,159 & 265,877 & TSFresh \\
 & & 0:15 & 2,537 & 4,781 & Catabra \\
\cmidrule(r){2-6}
 & \multirow{3}{*}{100} & \textbf{0:05} & \textbf{718} & \textbf{1,167} &  MEDS-Tab \\
 & & 5:09 & 217,477 & 659,735 & TSFresh \\
 & & 3:17 & 14,319 & 28,342 &  Catabra \\
\cmidrule(r){2-6}
 & 500 & \textbf{0:16} & \textbf{1,410} & \textbf{3,539} & MEDS-Tab \\
\bottomrule
\end{tabular}
\end{table}

For comprehensive results, computational efficiency comparisons, and further details on model performance across various datasets, see Table \ref{tab:combined_performance}. This repository serves as a resource for ongoing updates, additional analyses, and tutorials to enhance community engagement and facilitate continuous benchmarking efforts and collaborations on generalizable MEDS event stream methods. The repository can be accessed at \url{https://github.com/mmcdermott/MEDS_Tabular_AutoML/}.

\subsection{Limitations and Future Roadmap}

MEDS-Tab opens up new research avenues by providing an efficient tabularization and scalable featurization framework to access the functionally infinite feature space of EHR data. However, there are areas where further enhancements are anticipated. Future developments will focus on incorporating additional aggregation functions that are tailored to time-related features, along with implementing various windowing strategies, such as windows that are event-bound (i.e. windows of various lengths that are defined with start and end points at specific events rather than over predefined, specific window lengths) to offer more contextually relevant data snapshots. Further improvements are planned to optimize the data storage and data loading processes to improve performance and scalability. Optimizing the computation of aggregations to reduce time and resource overhead is another key area for development. Additionally, integrating more pipeline operations, such as further dimensionality reduction and imputation methods, will further bolster the framework's capability to handle diverse and complex datasets. These enhancements will not only address current limitations but also broaden the applicability of MEDS-Tab across different medical data analysis scenarios, paving the way for more robust and versatile healthcare analytical tools.

\subsection{Addressing Current Limitations in the Field}
MEDS-Tab directly addresses several key limitations observed in current healthcare ML research practices:

Reproducibility: By providing a standardized approach to feature extraction and aggregation, MEDS-Tab significantly enhances the reproducibility of tabular baseline modeling. Researchers using MEDS-Tab only need to specify the configured inputs, including allowed codes, window sizes, and feature aggregation mechanisms to reproduce results.

Scalability: Unlike manual approaches that often necessitate subsampling large datasets, MEDS-Tab's efficient design allows researchers to utilize larger portions of datasets without manual subsampling. This enables more comprehensive analyses and potentially more robust models.

Feature Handling: MEDS-Tab's approach to feature extraction mitigates the risk of overlooking potentially important low-frequency features, which can be crucial in medical contexts where rare events may have high predictive value.

Standardization: MEDS-Tab offers a consistent methodology for tabularization and modeling across different EHR datasets, facilitating more reliable comparisons between studies and datasets.

By addressing these limitations, MEDS-Tab not only simplifies the process of generating baseline models but also elevates the overall quality and reliability of research in the field of healthcare ML.

\section{Conclusion} \label{sec:conclusion}
It is important to emphasize that MEDS-Tab itself is not a baseline model, but rather a powerful tool designed to enable researchers to efficiently produce robust, reproducible baseline models with minimal effort. By providing this standardized, scalable framework, MEDS-Tab contributes to advancing the state of healthcare ML research as a whole, promoting more consistent, comparable, and reliable studies across diverse datasets and clinical settings.

In this work, we have introduced MEDS-Tab, an efficient and scalable framework that addresses critical challenges in the tabularization, featurization, and baseline modeling of EHR data. By integrating the MEDS format and its ecosystem, MEDS-Tab enhances the reproducibility, robustness, and reliability of health data analysis and reduces the computational burden often imposed by large-scale data. MEDS-Tab processes data in sparse matrix representations and employs data-sharding techniques that combine to create a scalable solution for processing and modeling large EHR data. Furthermore, MEDS-Tab provides a robust and flexible AutoML pipeline powered by Optuna to train high-performing XGBoost models to create competitive, comparable baselines in the medical space. With all these advancements, MEDS-Tab provides the medical research community with a powerful tabularization, featurization, and AutoML tool that promotes more effective and efficient data-driven healthcare solutions.

\begin{ack}
MBAM gratefully acknowledges support from a Berkowitz Postdoctoral Fellowship at Harvard Medical School.

 We thank the creators of the MIMIC-IV and eICU datasets for their tireless work in designing such large-scale collections of medical data and for making them publicly available.
\end{ack}

\bibliographystyle{plain}
\bibliography{references}

\begin{thebibliography}{10}

\bibitem{akiba2019optuna}
Takuya Akiba, Shotaro Sano, Toshihiko Yanase, Takeru Ohta, and Masanori Koyama.
\newblock Optuna: A next-generation hyperparameter optimization framework.
\newblock In {\em Proceedings of the 25th ACM SIGKDD international conference
  on knowledge discovery \& data mining}, pages 2623--2631, 2019.

\bibitem{alnegheimish2020cardea}
Sarah Alnegheimish, Najat Alrashed, Faisal Aleissa, Shahad Althobaiti, Dongyu
  Liu, Mansour Alsaleh, and Kalyan Veeramachaneni.
\newblock Cardea: An open automated machine learning framework for electronic
  health records.
\newblock In {\em 2020 IEEE 7th International Conference on Data Science and
  Advanced Analytics (DSAA)}, pages 536--545. IEEE, 2020.

\bibitem{MEDS}
Bert Arnrich, Edward Choi, Jason~Alan Fries, Matthew~B.A. McDermott, Jungwoo
  Oh, Tom Pollard, Nigam Shah, Ethan Steinberg, Michael Wornow, and Robin
  van~de Water.
\newblock Medical event data standard ({MEDS}): Facilitating machine learning
  for health.
\newblock In {\em ICLR 2024 Workshop on Learning from Time Series For Health},
  2024.

\bibitem{aydin2021performance}
Zeliha~Ergul Aydin and Zehra~Kamisli Ozturk.
\newblock Performance analysis of xgboost classifier with missing data.
\newblock In {\em 1st Int. Conf. Comput. Mach. Intell., no}, 2021.

\bibitem{bergstra2015hyperopt}
James Bergstra, Brent Komer, Chris Eliasmith, Dan Yamins, and David~D Cox.
\newblock Hyperopt: a python library for model selection and hyperparameter
  optimization.
\newblock {\em Computational Science \& Discovery}, 8(1):014008, 2015.

\bibitem{borisov2022deep}
Vadim Borisov, Tobias Leemann, Kathrin Se{\ss}ler, Johannes Haug, Martin
  Pawelczyk, and Gjergji Kasneci.
\newblock Deep neural networks and tabular data: A survey.
\newblock {\em IEEE Transactions on Neural Networks and Learning Systems},
  2022.

\bibitem{chen2016xgboost}
Tianqi Chen and Carlos Guestrin.
\newblock Xgboost: A scalable tree boosting system.
\newblock In {\em Proceedings of the 22nd acm sigkdd international conference
  on knowledge discovery and data mining}, pages 785--794, 2016.

\bibitem{christ2018time}
Maximilian Christ, Nils Braun, Julius Neuffer, and Andreas~W Kempa-Liehr.
\newblock Time series feature extraction on basis of scalable hypothesis tests
  (tsfresh--a python package).
\newblock {\em Neurocomputing}, 307:72--77, 2018.

\bibitem{de2023choice}
Lucas~BV de~Amorim, George~DC Cavalcanti, and Rafael~MO Cruz.
\newblock The choice of scaling technique matters for classification
  performance.
\newblock {\em Applied Soft Computing}, 133:109924, 2023.

\bibitem{1_MLHC2023}
Kaivalya Deshpande, Madalina Fiterau, Shalmali Joshi, Zachary Lipton, Rajesh
  Ranganath, Iñigo Urteaga, and Serene Yeung, editors.
\newblock {\em Proceedings of the 8th Machine Learning for Healthcare
  Conference}, volume 219 of {\em Proceedings of Machine Learning Research}.
  PMLR.

\bibitem{4_pmlr-v219-elhussein23a}
Ahmed Elhussein and Gamze G\"ursoy.
\newblock Privacy-preserving patient clustering for personalized federated
  learnings.
\newblock In Kaivalya Deshpande, Madalina Fiterau, Shalmali Joshi, Zachary
  Lipton, Rajesh Ranganath, Iñigo Urteaga, and Serene Yeung, editors, {\em
  Proceedings of the 8th Machine Learning for Healthcare Conference}, volume
  219 of {\em Proceedings of Machine Learning Research}, pages 150--166. PMLR,
  11--12 Aug 2023.

\bibitem{erickson2020autogluon}
Nick Erickson, Jonas Mueller, Alexander Shirkov, Hang Zhang, Pedro Larroy,
  Mu~Li, and Alexander Smola.
\newblock Autogluon-tabular: Robust and accurate automl for structured data.
\newblock {\em arXiv preprint arXiv:2003.06505}, 2020.

\bibitem{feurer2020auto}
Matthias Feurer, Katharina Eggensperger, Stefan Falkner, Marius Lindauer, and
  Frank Hutter.
\newblock Auto-sklearn 2.0: The next generation.
\newblock {\em arXiv preprint arXiv:2007.04074}, 24:8, 2020.

\bibitem{fu2022hint}
Tianfan Fu, Kexin Huang, Cao Xiao, Lucas~M Glass, and Jimeng Sun.
\newblock Hint: Hierarchical interaction network for clinical-trial-outcome
  predictions.
\newblock {\em Patterns}, 3(4), 2022.

\bibitem{harutyunyan2019multitask}
Hrayr Harutyunyan, Hrant Khachatrian, David~C Kale, Greg Ver~Steeg, and Aram
  Galstyan.
\newblock Multitask learning and benchmarking with clinical time series data.
\newblock {\em Scientific data}, 6(1):96, 2019.

\bibitem{hastie2009elements}
Trevor Hastie.
\newblock The elements of statistical learning: data mining, inference, and
  prediction, 2009.

\bibitem{2_pmlr-v225-hegselmann23a}
Stefan Hegselmann, Antonio Parziale, Divya Shanmugam, Shengpu Tang, Kristen
  Severson, Mercy~Nyamewaa Asiedu, Serina Chang, Bonaventure F.~P. Dossou, Qian
  Huang, Fahad Kamran, Haoran Zhang, Sujay Nagaraj, Luis Oala, Shan Xu,
  Chinasa~T. Okolo, Helen Zhou, Jessica Dafflon, Caleb Ellington, Sarah
  Jabbour, Hyewon Jeong, Harry~Reyes Nieva, Yuzhe Yang, Ghada Zamzmi, Vishwali
  Mhasawade, Van Truong, Payal Chandak, Matthew Lee, Peniel Argaw, Kyle Heuton,
  Harvineet Singh, and Thomas Hartvigsen.
\newblock Machine learning for health (ml4h) 2023.
\newblock In Stefan Hegselmann, Antonio Parziale, Divya Shanmugam, Shengpu
  Tang, Mercy~Nyamewaa Asiedu, Serina Chang, Tom Hartvigsen, and Harvineet
  Singh, editors, {\em Proceedings of the 3rd Machine Learning for Health
  Symposium}, volume 225 of {\em Proceedings of Machine Learning Research},
  pages 1--12. PMLR, 10 Dec 2023.

\bibitem{5_pmlr-v219-ho23a}
Danliang Ho and Mehul Motani.
\newblock Multi-view modelling of longitudinal health data for improved
  prognostication of colorectal cancer recurrence.
\newblock In Kaivalya Deshpande, Madalina Fiterau, Shalmali Joshi, Zachary
  Lipton, Rajesh Ranganath, Iñigo Urteaga, and Serene Yeung, editors, {\em
  Proceedings of the 8th Machine Learning for Healthcare Conference}, volume
  219 of {\em Proceedings of Machine Learning Research}, pages 265--284. PMLR,
  11--12 Aug 2023.

\bibitem{jarrett2022hyperimpute}
Daniel Jarrett, Bogdan~C Cebere, Tennison Liu, Alicia Curth, and Mihaela
  van~der Schaar.
\newblock Hyperimpute: Generalized iterative imputation with automatic model
  selection.
\newblock In {\em International Conference on Machine Learning}, pages
  9916--9937. PMLR, 2022.

\bibitem{jarrett2023clairvoyance}
Daniel Jarrett, Jinsung Yoon, Ioana Bica, Zhaozhi Qian, Ari Ercole, and Mihaela
  van~der Schaar.
\newblock Clairvoyance: A pipeline toolkit for medical time series.
\newblock {\em arXiv preprint arXiv:2310.18688}, 2023.

\bibitem{jensen2012mining}
Peter~B Jensen, Lars~J Jensen, and S{\o}ren Brunak.
\newblock Mining electronic health records: towards better research
  applications and clinical care.
\newblock {\em Nature Reviews Genetics}, 13(6):395--405, 2012.

\bibitem{ebcl}
Hyewon Jeong, Nassim Oufattole, Aparna Balagopalan, Matthew Mcdermott, Payal
  Chandak, Marzyeh Ghassemi, and Collin Stultz.
\newblock Event-based contrastive learning for medical time series.
\newblock {\em arXiv preprint arXiv:2312.10308}, 2023.

\bibitem{johnson2020mimic}
Alistair Johnson, Lucas Bulgarelli, Tom Pollard, Steven Horng, Leo~Anthony
  Celi, and Roger Mark.
\newblock Mimic-iv.
\newblock {\em PhysioNet. Available online at: https://physionet.
  org/content/mimiciv/1.0/(accessed August 23, 2021)}, pages 49--55, 2020.

\bibitem{mimiciv}
Alistair~EW Johnson, Lucas Bulgarelli, Lu~Shen, Alvin Gayles, Ayad Shammout,
  Steven Horng, Tom~J Pollard, Sicheng Hao, Benjamin Moody, Brian Gow, et~al.
\newblock Mimic-iv, a freely accessible electronic health record dataset.
\newblock {\em Scientific data}, 10(1):1, 2023.

\bibitem{6_pmlr-v225-king23a}
Ryan King, Tianbao Yang, and Bobak~J. Mortazavi.
\newblock Multimodal pretraining of medical time series and notes.
\newblock In Stefan Hegselmann, Antonio Parziale, Divya Shanmugam, Shengpu
  Tang, Mercy~Nyamewaa Asiedu, Serina Chang, Tom Hartvigsen, and Harvineet
  Singh, editors, {\em Proceedings of the 3rd Machine Learning for Health
  Symposium}, volume 225 of {\em Proceedings of Machine Learning Research},
  pages 244--255. PMLR, 10 Dec 2023.

\bibitem{omoplearn}
Rohan Kodialam, Rebecca Boiarsky, Justin Lim, Aditya Sai, Neil Dixit, and David
  Sontag.
\newblock Deep contextual clinical prediction with reverse distillation.
\newblock In {\em Proceedings of the AAAI Conference on Artificial
  Intelligence}, volume~35, pages 249--258, 2021.

\bibitem{7_pmlr-v225-kuznetsova23a}
Rita Kuznetsova, Aliz\'ee Pace, Manuel Burger, Hugo Y\`eche, and Gunnar
  R\"atsch.
\newblock On the importance of step-wise embeddings for heterogeneous clinical
  time-series.
\newblock In Stefan Hegselmann, Antonio Parziale, Divya Shanmugam, Shengpu
  Tang, Mercy~Nyamewaa Asiedu, Serina Chang, Tom Hartvigsen, and Harvineet
  Singh, editors, {\em Proceedings of the 3rd Machine Learning for Health
  Symposium}, volume 225 of {\em Proceedings of Machine Learning Research},
  pages 268--291. PMLR, 10 Dec 2023.

\bibitem{8_pmlr-v219-labach23a}
Alex Labach, Aslesha Pokhrel, Xiao~Shi Huang, Saba Zuberi, Seung~Eun Yi,
  Maksims Volkovs, Tomi Poutanen, and Rahul~G. Krishnan.
\newblock Duett: Dual event time transformer for electronic health records.
\newblock In Kaivalya Deshpande, Madalina Fiterau, Shalmali Joshi, Zachary
  Lipton, Rajesh Ranganath, Iñigo Urteaga, and Serene Yeung, editors, {\em
  Proceedings of the 8th Machine Learning for Healthcare Conference}, volume
  219 of {\em Proceedings of Machine Learning Research}, pages 403--422. PMLR,
  11--12 Aug 2023.

\bibitem{li2021imputation}
Jiang Li, Xiaowei~S Yan, Durgesh Chaudhary, Venkatesh Avula, Satish Mudiganti,
  Hannah Husby, Shima Shahjouei, Ardavan Afshar, Walter~F Stewart, Mohammed
  Yeasin, et~al.
\newblock Imputation of missing values for electronic health record laboratory
  data.
\newblock {\em NPJ digital medicine}, 4(1):147, 2021.

\bibitem{li2020learning}
Steven Cheng-Xian Li and Benjamin Marlin.
\newblock Learning from irregularly-sampled time series: A missing data
  perspective.
\newblock In {\em International Conference on Machine Learning}, pages
  5937--5946. PMLR, 2020.

\bibitem{loning2019sktime}
Markus L{\"o}ning, Anthony Bagnall, Sajaysurya Ganesh, Viktor Kazakov, Jason
  Lines, and Franz~J Kir{\'a}ly.
\newblock sktime: A unified interface for machine learning with time series.
\newblock {\em arXiv preprint arXiv:1909.07872}, 2019.

\bibitem{maletzky2023catabra}
Alexander Maletzky, Sophie Kaltenleithner, Philipp Moser, and Michael
  Giretzlehner.
\newblock Catabra: Efficient analysis and predictive modeling of tabular data.
\newblock In {\em IFIP International Conference on Artificial Intelligence
  Applications and Innovations}, pages 57--68. Springer, 2023.

\bibitem{mcdermott2020comprehensive}
Matthew McDermott, Bret Nestor, Evan Kim, Wancong Zhang, Anna Goldenberg, Peter
  Szolovits, and Marzyeh Ghassemi.
\newblock A comprehensive evaluation of multi-task learning and multi-task
  pre-training on ehr time-series data.
\newblock {\em arXiv preprint arXiv:2007.10185}, 2020.

\bibitem{nguyen2021clinical}
A~Tuan Nguyen, Hyewon Jeong, Eunho Yang, and Sung~Ju Hwang.
\newblock Clinical risk prediction with temporal probabilistic asymmetric
  multi-task learning.
\newblock In {\em Proceedings of the AAAI Conference on Artificial
  Intelligence}, volume~35, pages 9081--9091, 2021.

\bibitem{9_pmlr-v225-noroozizadeh23a}
Shahriar Noroozizadeh, Jeremy~C. Weiss, and George~H. Chen.
\newblock Temporal supervised contrastive learning for modeling patient risk
  progression.
\newblock In Stefan Hegselmann, Antonio Parziale, Divya Shanmugam, Shengpu
  Tang, Mercy~Nyamewaa Asiedu, Serina Chang, Tom Hartvigsen, and Harvineet
  Singh, editors, {\em Proceedings of the 3rd Machine Learning for Health
  Symposium}, volume 225 of {\em Proceedings of Machine Learning Research},
  pages 403--427. PMLR, 10 Dec 2023.

\bibitem{pang2022establishment}
Ke~Pang, Liang Li, Wen Ouyang, Xing Liu, and Yongzhong Tang.
\newblock Establishment of icu mortality risk prediction models with machine
  learning algorithm using mimic-iv database.
\newblock {\em Diagnostics}, 12(5):1068, 2022.

\bibitem{pedregosa2011scikit}
Fabian Pedregosa, Ga{\"e}l Varoquaux, Alexandre Gramfort, Vincent Michel,
  Bertrand Thirion, Olivier Grisel, Mathieu Blondel, Peter Prettenhofer, Ron
  Weiss, Vincent Dubourg, et~al.
\newblock Scikit-learn: Machine learning in python.
\newblock {\em the Journal of machine Learning research}, 12:2825--2830, 2011.

\bibitem{3_pmlr-v248-pollard24a}
Tom Pollard, Edward Choi, Pankhuri Singhal, Michael Hughes, Elena Sizikova,
  Bobak Mortazavi, Irene Chen, Fei Wang, Tasmie Sarker, Matthew McDermott, and
  Marzyeh Ghassemi.
\newblock Conference on health, inference, and learning (chil) 2024.
\newblock In Tom Pollard, Edward Choi, Pankhuri Singhal, Michael Hughes, Elena
  Sizikova, Bobak Mortazavi, Irene Chen, Fei Wang, Tasmie Sarker, Matthew
  McDermott, and Marzyeh Ghassemi, editors, {\em Proceedings of the fifth
  Conference on Health, Inference, and Learning}, volume 248 of {\em
  Proceedings of Machine Learning Research}, pages 1--6. PMLR, 27--28 Jun 2024.

\bibitem{pollard2018eicu}
Tom~J Pollard, Alistair E~W Johnson, Jesse~D Raffa, Leo~A Celi, Roger~G Mark,
  and Omar Badawi.
\newblock The {eICU Collaborative Research Database}, a freely available
  multi-center database for critical care research.
\newblock {\em Scientific data}, 5(1):1--13, 2018.

\bibitem{rashidi2021machine}
Hooman~H Rashidi, Nam Tran, Samer Albahra, and Luke~T Dang.
\newblock Machine learning in health care and laboratory medicine: General
  overview of supervised learning and auto-ml.
\newblock {\em International Journal of Laboratory Hematology}, 43:15--22,
  2021.

\bibitem{10_pmlr-v225-ren23a}
Yifei Ren, Jian Lou, Li~Xiong, Joyce~C Ho, Xiaoqian Jiang, and
  Sivasubramanium~Venkatraman Bhavani.
\newblock Multipar: Supervised irregular tensor factorization with multi-task
  learning for computational phenotyping.
\newblock In Stefan Hegselmann, Antonio Parziale, Divya Shanmugam, Shengpu
  Tang, Mercy~Nyamewaa Asiedu, Serina Chang, Tom Hartvigsen, and Harvineet
  Singh, editors, {\em Proceedings of the 3rd Machine Learning for Health
  Symposium}, volume 225 of {\em Proceedings of Machine Learning Research},
  pages 498--511. PMLR, 10 Dec 2023.

\bibitem{rusdah2020xgboost}
Deandra~Aulia Rusdah and Hendri Murfi.
\newblock Xgboost in handling missing values for life insurance risk
  prediction.
\newblock {\em SN Applied Sciences}, 2(8):1336, 2020.

\bibitem{saveliev2023temporai}
Evgeny~S Saveliev and Mihaela van~der Schaar.
\newblock Temporai: Facilitating machine learning innovation in time domain
  tasks for medicine.
\newblock {\em arXiv preprint arXiv:2301.12260}, 2023.

\bibitem{tabular_baselines}
Ravid Shwartz-Ziv and Amitai Armon.
\newblock Tabular data: Deep learning is not all you need.
\newblock In {\em 8th ICML Workshop on Automated Machine Learning (AutoML)},
  2021.

\bibitem{singh2022automated}
Vivek~Kumar Singh and Kailash Joshi.
\newblock Automated machine learning (automl): an overview of opportunities for
  application and research.
\newblock {\em Journal of Information Technology Case and Application
  Research}, 24(2):75--85, 2022.

\bibitem{smith2020machine}
Micah~J Smith, Carles Sala, James~Max Kanter, and Kalyan Veeramachaneni.
\newblock The machine learning bazaar: Harnessing the ml ecosystem for
  effective system development.
\newblock In {\em Proceedings of the 2020 ACM SIGMOD International Conference
  on Management of Data}, pages 785--800, 2020.

\bibitem{sun2022effects}
Wen Sun, Yang Yan, Shidong Hu, Boyan Liu, Shuying Wang, Wenli Yu, and Songyan
  Li.
\newblock The effects of midazolam or propofol plus fentanyl on icu mortality:
  a retrospective study based on the mimic-iv database.
\newblock {\em Annals of Translational Medicine}, 10(4), 2022.

\bibitem{sun2023prediction}
Yiwu Sun, Zhaoyi He, Jie Ren, and Yifan Wu.
\newblock Prediction model of in-hospital mortality in intensive care unit
  patients with cardiac arrest: a retrospective analysis of mimic-iv database
  based on machine learning.
\newblock {\em BMC anesthesiology}, 23(1):178, 2023.

\bibitem{tipirneni2022self}
Sindhu Tipirneni and Chandan~K Reddy.
\newblock Self-supervised transformer for sparse and irregularly sampled
  multivariate clinical time-series.
\newblock {\em ACM Transactions on Knowledge Discovery from Data (TKDD)},
  16(6):1--17, 2022.

\bibitem{11_pmlr-v225-van-ness23a}
Mike Van~Ness, Tomas Bosschieter, Natasha Din, Andrew Ambrosy, Alexander
  Sandhu, and Madeleine Udell.
\newblock Interpretable survival analysis for heart failure risk prediction.
\newblock In Stefan Hegselmann, Antonio Parziale, Divya Shanmugam, Shengpu
  Tang, Mercy~Nyamewaa Asiedu, Serina Chang, Tom Hartvigsen, and Harvineet
  Singh, editors, {\em Proceedings of the 3rd Machine Learning for Health
  Symposium}, volume 225 of {\em Proceedings of Machine Learning Research},
  pages 574--593. PMLR, 10 Dec 2023.

\bibitem{weiskopf2013defining}
Nicole~G Weiskopf, George Hripcsak, Sushmita Swaminathan, and Chunhua Weng.
\newblock Defining and measuring completeness of electronic health records for
  secondary use.
\newblock {\em Journal of biomedical informatics}, 46(5):830--836, 2013.

\bibitem{wells2013strategies}
Brian~J Wells, Kevin~M Chagin, Amy~S Nowacki, and Michael~W Kattan.
\newblock Strategies for handling missing data in electronic health record
  derived data.
\newblock {\em Egems}, 1(3), 2013.

\bibitem{12_pmlr-v248-xu24a}
Ran Xu, Yiwen Lu, Chang Liu, Yong Chen, Yan Sun, Xiao Hu, Joyce~C Ho, and Carl
  Yang.
\newblock From basic to extra features: Hypergraph transformer
  pretrain-then-finetuning for balanced clinical predictions on ehr.
\newblock In Tom Pollard, Edward Choi, Pankhuri Singhal, Michael Hughes, Elena
  Sizikova, Bobak Mortazavi, Irene Chen, Fei Wang, Tasmie Sarker, Matthew
  McDermott, and Marzyeh Ghassemi, editors, {\em Proceedings of the fifth
  Conference on Health, Inference, and Learning}, volume 248 of {\em
  Proceedings of Machine Learning Research}, pages 182--197. PMLR, 27--28 Jun
  2024.

\bibitem{13_pmlr-v225-xu23a}
Yanbo Xu, Shangqing Xu, Manav Ramprassad, Alexey Tumanov, and Chao Zhang.
\newblock Transehr: Self-supervised transformer for clinical time series data.
\newblock In Stefan Hegselmann, Antonio Parziale, Divya Shanmugam, Shengpu
  Tang, Mercy~Nyamewaa Asiedu, Serina Chang, Tom Hartvigsen, and Harvineet
  Singh, editors, {\em Proceedings of the 3rd Machine Learning for Health
  Symposium}, volume 225 of {\em Proceedings of Machine Learning Research},
  pages 623--635. PMLR, 10 Dec 2023.

\bibitem{14_pmlr-v248-yeche24a}
Hugo Y\`eche, Manuel Burger, Dinara Veshchezerova, and Gunnar Ratsch.
\newblock Dynamic survival analysis for early event prediction.
\newblock In Tom Pollard, Edward Choi, Pankhuri Singhal, Michael Hughes, Elena
  Sizikova, Bobak Mortazavi, Irene Chen, Fei Wang, Tasmie Sarker, Matthew
  McDermott, and Marzyeh Ghassemi, editors, {\em Proceedings of the fifth
  Conference on Health, Inference, and Learning}, volume 248 of {\em
  Proceedings of Machine Learning Research}, pages 540--557. PMLR, 27--28 Jun
  2024.

\bibitem{15_pmlr-v219-zhang23a}
Lida Zhang and Bobak~J. Mortazavi.
\newblock Semi-supervised meta-learning for multi-source heterogeneity in
  time-series data.
\newblock In Kaivalya Deshpande, Madalina Fiterau, Shalmali Joshi, Zachary
  Lipton, Rajesh Ranganath, Iñigo Urteaga, and Serene Yeung, editors, {\em
  Proceedings of the 8th Machine Learning for Healthcare Conference}, volume
  219 of {\em Proceedings of Machine Learning Research}, pages 923--941. PMLR,
  11--12 Aug 2023.

\bibitem{zhao2021pyhealth}
Yue Zhao, Zhi Qiao, Cao Xiao, Lucas Glass, and Jimeng Sun.
\newblock Pyhealth: A python library for health predictive models.
\newblock {\em arXiv preprint arXiv:2101.04209}, 2021.

\bibitem{zhou2017learning}
Chongyu Zhou, Yao Jia, Mehul Motani, and Jingwei Chew.
\newblock Learning deep representations from heterogeneous patient data for
  predictive diagnosis.
\newblock In {\em Proceedings of the 8th ACM International Conference on
  Bioinformatics, Computational Biology, and Health Informatics}, pages
  115--123, 2017.

\end{thebibliography}

\end{document}